\documentclass{article}
\usepackage{spconf,amsmath,graphicx,hyperref}
\usepackage{booktabs}
\usepackage{multirow}
\usepackage{subcaption}
\usepackage[table]{xcolor}
\usepackage{colortbl}

\title{SLOW-FAST MULTI-TEACHER ON-POLICY DISTILLATION \\
FOR CAPABILITY PRESERVATION}

\name{ Xiaofei Yin\textsuperscript{1,\dag}, Tong Chu\textsuperscript{1,\dag}, Jiyuan Fu\textsuperscript{2}, Jun Lan\textsuperscript{1}, Shuheng  Zhou\textsuperscript{1,*}, Huijia Zhu\textsuperscript{1,*}\thanks{*Corresponding author, \dag Contributed equally }}
\address{\textsuperscript{1}Ant Security Lab, Ant Group, Shanghai, China 
\textsuperscript{2}Fudan University, Shanghai, China}

\begin{document}
%
\maketitle
\begin{abstract}
Foundation multimodal large language models are designed to support a broad spectrum of capabilities across diverse domains. Multi-teacher on-policy distillation (MOPD) provides an effective framework for consolidating domain-specific expertise into a single student model. However, MOPD training gradually drives the student away from its initialization model, and general capabilities decline as the displacement grows, resulting in capability interference. A direct remedy is constraining the student toward its initialization, but this suppresses the acquisition of domain expertise as well. We propose Slow-Fast Multi-Teacher On-Policy Distillation (SF-MOPD), which couples a fast model, the current student updated directly by each teacher, with a slow model, an exponential moving average of the student. The slow model absorbs the learning signal gradually, serving as a moving capability reference that fuses the general foundation with confirmed domain expertise. For each teacher, SF-MOPD computes the teacher-induced update in log-probability space and removes only the component that pushes the fast model further away from the slow model, while retaining aligned and orthogonal components. Experiments across multiple model scales demonstrate that SF-MOPD effectively mitigates capability interference, enhances specialized multimodal capabilities, and reduces the average degradation on general-capability benchmarks, consistently outperforming vanilla MOPD.
\end{abstract}
\vspace{-0.1cm}

\begin{figure}[t]
    \centering
    \begin{subfigure}[t]{0.49\linewidth}
        \centering
        \includegraphics[width=\linewidth]{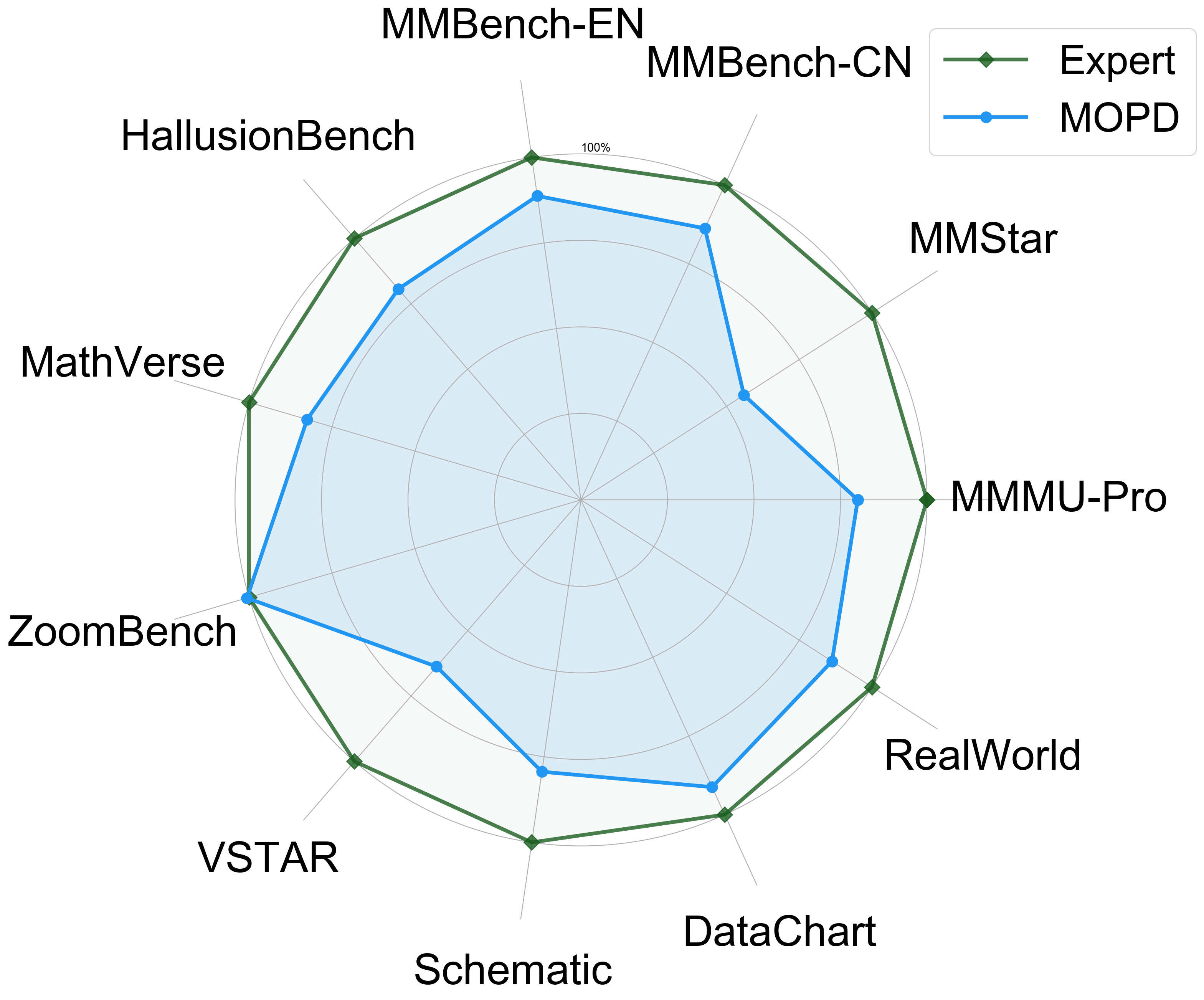}
    \end{subfigure}
    \hfill
    \begin{subfigure}[t]{0.49\linewidth}
        \centering
        \includegraphics[width=\linewidth]{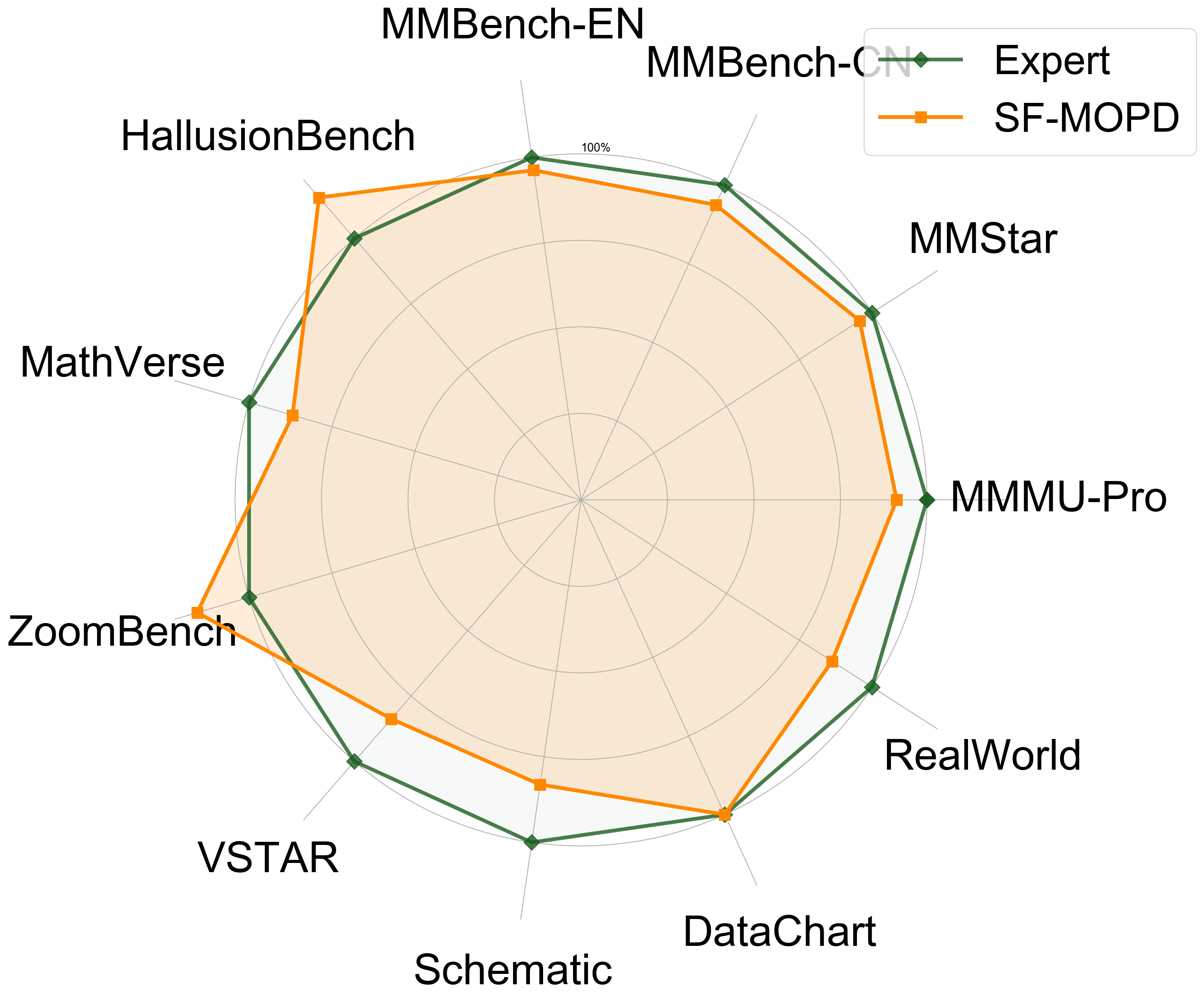}
    \end{subfigure}
    \vspace{-0.3cm}
    \caption{Comparison of capability absorption by MOPD and SF-MOPD across 11 benchmarks, with scores normalized separately for each dataset. The green line denotes the per-benchmark oracle reference obtained from the corresponding expert. SF-MOPD approaches the oracle reference on most benchmarks while better preserving general capabilities.}
    \label{fig:radar}
    \vspace{-0.8cm}
\end{figure}

\section{Introduction}
\label{sec:intro}

Multimodal large language models (MLLMs) have demonstrated strong capabilities across diverse domains~\cite{bai2025qwen3}, commonly developed through supervised fine-tuning and reinforcement learning~\cite{bai2025qwen3}. Multi-teacher on-policy distillation (MOPD)~\cite{ma2026mopd} has emerged as an effective approach to consolidating complementary expertise from multiple domain-specific experts into a single student, yet gradually drives the student away from its initialization, a phenomenon we refer to as \emph{capability interference}.

\begin{figure*}[t]
    \centering
    \includegraphics[width=1.0\linewidth]{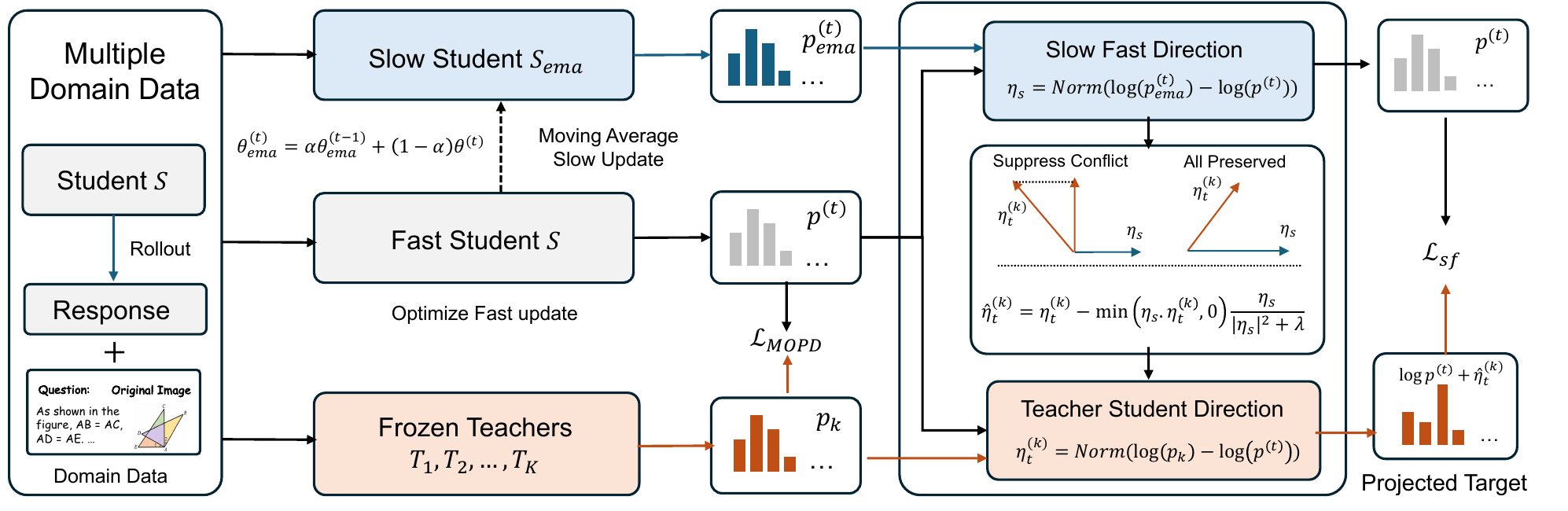}
    \vspace{-0.7cm}
    \caption{Overview of the SF-MOPD pipeline. Multi-domain data flows through a fast-slow student-teacher framework.}
    \label{fig:pipline}
    \vspace{-0.7cm}
\end{figure*}

A natural remedy is to anchor the student to its initialization via KL-regularized fine-tuning~\cite{shao2024deepseekmath}, but a fixed anchor cannot separate domain expertise acquisition from general capability erosion, as penalizing the deviation suppresses both. We instead propose \emph{Slow-Fast Multi-Teacher On-Policy Distillation} (SF-MOPD), which couples a \emph{fast} model, the current student updated directly by each teacher, with a \emph{slow} model, an exponential moving average (EMA) of the student that starts from the same initialization and absorbs learning signals gradually so that recurring directions survive while transient fluctuations are smoothed away, serving as a moving capability reference. For each on-policy sample, SF-MOPD removes only the component of each teacher-induced update that enlarges the slow-fast displacement, retaining aligned and orthogonal components that may encode complementary knowledge. The delivered model is the fast model, which retains the full learning signal from every teacher while being kept from drifting beyond its confirmed capability state.

SF-MOPD introduces minimal overhead, requiring no teacher-specific gradient storage or additional trainable parameters, and extensive experiments across multiple model scales and specialized multimodal domains show that it consistently enhances expert-specific capabilities while better preserving general multimodal competence, outperforming vanilla MOPD. Our main contributions are as follows:
(1)~A slow-fast capability reference built on an EMA of the student, starting from the initialization and gradually absorbing confirmed domain expertise.
(2)~A per-teacher projection mechanism that removes only the update component enlarging the slow-fast displacement while retaining aligned and orthogonal components.
(3)~Extensive validation across multiple model scales and specialized multimodal domains, showing a consistent balance between expert capability integration and general capability preservation.
\section{METHOD}

\subsection{Preliminary}

\noindent Multi-Teacher On-Policy Distillation (MOPD) extends standard knowledge distillation by incorporating multiple teacher models to transfer complementary domain-specific capabilities into a single student model. Let $\mathcal{T} = \{T_1, \ldots, T_K\}$ denote the set of $K$ teacher models and $S$ the student model. Given an instruction $q$, the student model generates a response $r$ by sampling from its policy at the current training step $t$. The instruction and response are concatenated to form the input sequence $x = [q; r]$, which is then fed to both the student and each teacher to obtain logits:
\begin{equation}
    z^{(t)} = S(x), \qquad z_k = T_k(x), \quad k = 1, \ldots, K.
\end{equation}
The standard MOPD objective combines the distillation losses from all teachers:
\begin{equation}
    \mathcal{L}_{\text{MOPD}} =
    \sum_{k=1}^{K}
    \mathcal{L}_{\text{JSD}}\!\left(z^{(t)},\, z_k\right),
\end{equation}
where $\mathcal{L}_{\text{JSD}}$ is the Jensen-Shannon divergence between the predictive distributions of the student and the teacher:
\begin{equation}
\mathcal{L}_{\mathrm{JSD}}\!\left(z^{(t)}, z_k\right)
=
\mathrm{JSD}\!\left(
\sigma(z^{(t)})
\,\middle\|\,
\operatorname{sg}\!\left[\sigma(z_k)\right]
\right),
\end{equation}
where $\operatorname{sg}$ means stop-gradient and $\sigma$ denotes the softmax function. However, MOPD training gradually drives the student away from its initialization model, and general capabilities decline as the displacement grows. We refer to this phenomenon as capability interference~\cite{shao2024deepseekmath}.

\subsection{Slow-Fast Capability Reference}

As shown in Fig.~\ref{fig:pipline}, SF-MOPD maintains two models with complementary update rates. The \emph{fast} model is the current student, updated directly by each teacher on its own samples to rapidly acquire new expertise. The \emph{slow} model $S_{\mathrm{ema}}$ is maintained via exponential moving average (EMA) over the student parameters:
\begin{equation}
    \theta_{\mathrm{ema}}^{(t)}
    =
    \alpha\theta_{\mathrm{ema}}^{(t-1)}
    +
    (1-\alpha)\theta^{(t)},
\end{equation}
where $\alpha$ is the smoothing coefficient. Starting from the same initialization, the slow model evolves through smooth parameter fusion and stays closer to the initialization, where general capabilities are best preserved. By retaining recurring update directions and smoothing transient fluctuations, it gradually absorbs domain expertise on this general foundation. Consequently, keeping the fast model close to the slow model encourages the preservation of general capabilities throughout learning. We define the \emph{slow-fast constraint direction} as the difference between the log-probabilities of the slow and fast models:
\begin{equation}
    \eta_s =
    \log\sigma(z_{\mathrm{ema}}^{(t)})
    -
    \log\sigma(z^{(t)}),
\end{equation}
where \emph{CenterNorm} subtracts the mean from each log-probability vector to normalize $\eta_s$, capturing relative token-level adjustments rather than global shifts. Unlike a fixed direction toward the initialization logits $z^{(0)}$, this direction evolves with training and remains consistent with the expertise already absorbed.

\subsection{Slow-Fast Projection}

\noindent For each teacher $k$, we define the teacher-induced update direction as:
\begin{equation}
    \eta_t^{(k)} =
    \log\sigma(z_k)
    -
    \log\sigma(z^{(t)}),
\end{equation}
the terms $\eta_t^{(k)}$ are normalized by \emph{CenterNorm}. We compute its cosine similarity with the slow-fast constraint direction $\eta_s$:
\begin{equation}
    \rho_k =
    \frac{\eta_s\cdot\eta_t^{(k)}}
    {\|\eta_s\|\|\eta_t^{(k)}\|+\epsilon},
\end{equation}
where a negative $\rho_k$ indicates that the update induced by teacher $k$ pushes the fast model further away from the slow model. We isolate this opposing component via projection:
\begin{equation}
    \hat{\eta}_t^{(k)} =
    \eta_t^{(k)}-
    \min(\eta_s\cdot\eta_t^{(k)},0)\cdot
    \frac{\eta_s}{\|\eta_s\|^2+\lambda},
\end{equation}
where $\lambda$ is a ridge term for numerical stability. The projection removes the opposing component only when it exists, retaining aligned and orthogonal components that may encode complementary knowledge. The fast model is then trained to minimize:
\begin{equation}
    \mathcal{L}_{sf}
    =
    \sum_{k=1}^{K} \mathrm{JSD}\!\left(
    \sigma(z^{(t)})
    \;\Big\|\;
    \operatorname{sg}\!\left[\sigma\!\left(
    \log\sigma(z^{(t)})+\hat{\eta}_t^{(k)}
    \right)\right]
    \right),
\end{equation}
For numerical stability, we optionally clip $\hat{\eta}_t^{(k)}$ by a maximum norm $\delta$ and a per-element bound $c$, followed by re-centering. Since $\eta_s$ is derived entirely from the slow and fast models, the projection requires only parameter smoothing, with no teacher-specific gradient storage or additional trainable parameters.

\subsection{Overall Objective}

\noindent We retain the original MOPD loss as a regularization term, yielding the overall SF-MOPD objective:
\begin{equation}
    \mathcal{L}_{\mathrm{SF\text{-}MOPD}}
    =
    \mathcal{L}_{sf}
    +
    \beta \, \mathcal{L}_{\mathrm{MOPD}},
\end{equation}
where $\beta$ is a balancing coefficient. $\mathcal{L}_{\mathrm{MOPD}}$ provides unmodified supervision from all teachers, while $\mathcal{L}_{sf}$ corrects each teacher-induced update along the slow-fast constraint direction. Together, the two terms allow the fast model to absorb new domain-specific capabilities while remaining close to the capability state confirmed by the slow model.

\begin{table*}[htbp!]
\centering
\vspace{-0.8cm}
\caption{Comparison with state-of-the-art methods on General and Domain-Specific benchmarks. The best result is shown in bold, and the second-best result is underlined. All distillation methods are reproduced on our benchmarks under the same training data and budget. For each benchmark, we report the Expert result of the corresponding expert as an oracle reference.}
\label{tab:main_results}
\vspace{-0.3cm}
\setlength{\tabcolsep}{5pt}
{\Large
\resizebox{\textwidth}{!}{
\begin{tabular}{lcccccccccccccc}
\toprule
& \multicolumn{6}{c}{\textbf{General Capability}}
& \multicolumn{7}{c}{\textbf{Domain-Specific Capability}}
& \textbf{All}
\\
\cmidrule(lr){2-7} \cmidrule(lr){8-14} \cmidrule(lr){15-15}
\textbf{Method}
& \textbf{MMMU\_Pro}
& \textbf{MMStar}
& \textbf{MMBench-CN}
& \textbf{MMBench-EN}
& \textbf{HallusionBench}
& \textbf{Avg.}
& \textbf{MathVerse}
& \textbf{ZoomBench}
& \textbf{VSTAR}
& \textbf{Schematic}
& \textbf{DataChart}
& \textbf{RealWorld}
& \textbf{Avg.}
& \textbf{Avg.}
\\
\midrule
\multicolumn{15}{c}{\textit{Multi-Teacher Distillation: Qwen3-VL-2B-Instruct}} \\
\midrule
Qwen3-VL-2B-Instruct~\cite{bai2025qwen3} & \textbf{42.7} & \textbf{57.2} & \textbf{82.5} & \textbf{84.8} & \textbf{69.0} & \textbf{67.2} & 41.2 & 38.8 & 62.8 & \underline{31.5} & \underline{21.2} & \underline{35.1} & 38.4 & 51.5 \\
MOPD~\cite{ma2026mopd} & \textbf{42.7} & 52.2 & \underline{82.1} & \underline{83.7} & 66.3 & 65.4 & \underline{43.0} & \textbf{45.7} & \underline{66.0} & \textbf{43.0} & \textbf{27.9} & 34.2 & \underline{43.3} & \underline{53.3} \\
\rowcolor{gray!15} SF-MOPD (Ours) & \underline{42.3} & \underline{56.5} & 80.9 & 83.6 & \underline{68.9} & \underline{66.4} & \textbf{44.3} & \underline{45.3} & \textbf{67.5} & \textbf{43.0} & \textbf{27.9} & \textbf{36.4} & \textbf{44.1} & \textbf{54.2} \\
\midrule
\multicolumn{15}{c}{\textit{Multi-Teacher Distillation: Qwen3-VL-4B-Instruct}} \\
\midrule
Qwen3-VL-4B-Instruct~\cite{bai2025qwen3} & \textbf{55.3} & \textbf{69.4} & \textbf{88.3} & \underline{88.8} & \textbf{73.7} & \textbf{75.1} & \textbf{61.8} & 41.4 & \textbf{74.9} & 55.6 & 45.9 & 45.6 & 54.2 & \underline{63.7} \\
MOPD~\cite{ma2026mopd} & 51.9 & 60.7 & \underline{87.7} & 88.0 & 73.3 & 72.3 & 57.7 & \underline{48.0} & \underline{73.8} & \textbf{60.6} & \underline{49.3} & \underline{49.1} & \underline{56.4} & 63.6 \\
\rowcolor{gray!15} SF-MOPD (Ours) & \underline{54.6} & \underline{66.6} & 87.6 & \textbf{88.9} & \underline{73.5} & \underline{74.2} & \underline{59.3} & \textbf{48.8} & 73.3 & \underline{59.7} & \textbf{49.5} & \textbf{50.9} & \textbf{56.9} & \textbf{64.8} \\
\midrule
\multicolumn{15}{c}{\textit{Multi-Teacher Distillation: Qwen3-VL-8B-Instruct}} \\
\midrule
Qwen3-VL-8B-Instruct~\cite{bai2025qwen3} & \textbf{60.1} & \textbf{72.7} & \textbf{89.3} & \textbf{90.1} & \underline{74.0} & \textbf{77.2} & 62.9 & 39.1 & 75.9 & 63.1 & 49.8 & \underline{52.6} & 57.2 & \underline{66.3} \\
MOPD~\cite{ma2026mopd} & 56.9 & 63.5 & 87.1 & 88.3 & 70.9 & 73.3 & 62.5 & 46.6 & 75.9 & 64.0 & \underline{56.5} & 48.7 & 59.0 & 65.5 \\
MOPD~(Ref)~\cite{ma2026mopd} & 57.6 & 62.3 & 87.3 & 87.9 & 73.7 & 73.8 & \textbf{63.2} & 45.0 & 76.4 & 62.0 & 54.6 & \textbf{53.1} & \underline{59.1} & 65.7 \\
Open-MOPD~\cite{gao2026open} & 53.7 & 57.7 & 86.1 & 87.2 & 71.6 & 71.3 & 59.8 & 46.7 & 75.9 & \underline{64.3} & 54.3 & 48.7 & 58.3 & 64.2 \\
VAD-MOPD~\cite{zhang2026vad} & 52.2 & 62.9 & 86.6 & 87.9 & 65.9 & 71.1 & 57.5 & 44.9 & 78.0 & 63.4 & 53.1 & 46.9 & 57.3 & 63.6 \\
MeanTeacher~\cite{meanteacher} & 51.5 & 61.2 & 87.0 & 87.9 & \underline{74.0} & 72.3 & 59.9 & 45.2 & \underline{78.5} & 55.6 & 53.8 & 49.1 & 57.0 & 64.0 \\
OPD-V~\cite{bi2026opd} & 58.6 & 61.7 & 86.8 & 88.2 & 67.4 & 72.5 & \underline{63.0} & \underline{46.9} & 74.9 & 60.0 & 51.2 & 45.2 & 56.9 & 64.0 \\
\rowcolor{gray!15} SF-MOPD (Ours) & \underline{58.7} & \underline{68.8} & \underline{88.3} & \underline{89.5} & \textbf{76.5} & \underline{76.4} & \textbf{63.2} & \textbf{49.0} & \textbf{79.1} & \textbf{64.6} & \textbf{57.9} & 48.7 & \textbf{60.4} & \textbf{67.7} \\
\midrule
Expert~(Oracle Reference) & 60.1 & 72.7 & 89.3 & 90.1 & 74.0 & 77.2 & 65.3 & 46.5 & 81.7 & 67.3 & 57.9 & 50.9 & 61.6 & 68.7 \\
\bottomrule
\end{tabular}
}}
\vspace{-0.5cm}
\end{table*}

\section{EXPERIMENT}

\subsection{Experiment Setting}

\noindent \textbf{Datasets:}
DeepVision-103K~\cite{sun2026deepvision} is a diverse dataset spanning multiple visual domains. We partition it by domain capability to obtain three training sets: DataChart, RealWorld Item, and Schematic Diagram. To ensure sufficient heterogeneity for multi-teacher distillation, we additionally introduce Vision-OPD-6K~\cite{yuan2026vision} as the fine-grained visual perception set. For each domain, we randomly sample a subset of the corresponding training data. To measure domain-specific capabilities, we use MathVerse~\cite{zhang2024mathverse}, VSTAR~\cite{wu2024v} and ZoomBench~\cite{wei2026zooming} alongside the in-domain benchmarks. To measure general capability retention, we further include MMMU-Pro~\cite{yue2025mmmu}, MMBench-CN~\cite{liu2024mmbench}, MMBench-EN~\cite{liu2024mmbench}, MMStar~\cite{chen2024we}, and HallusionBench~\cite{guan2024hallusionbench} for evaluation.

\noindent \textbf{Setting:}
For the first three domains, we employ GSPO~\cite{gspo} to train the respective expert teacher models. For the perception expert, we follow the data construction methodology of Vision-OPD~\cite{yuan2026vision} and use the Vision-OPD-6K dataset as its training data. We use the Qwen3-VL-8B expert models as teachers and distill their knowledge into student models of 8B, 4B, and 2B scales, respectively. For evaluation, we first apply rule-based matching to extract model answers, and for cases where rule-based matching fails, we employ an LLM-as-judge strategy using Qwen3-VL-8B-Instruct to determine the correctness of the responses.

\noindent \textbf{Hyperparameters:}
We use the optimizer AdamW, learning rate $1\times10^{-5}$, batch size 64, and training one epoch for all compared methods at each model scale. The EMA smoothing coefficient is set to $\alpha=0.999$ for the all models. The coefficient of the original MOPD regularization term is $\beta=1.0$ for the 2B and 4B models and $\beta=0.1$ for the 8B model. For numerical stability, we use a ridge term of $\lambda=10^{-3}$ in the projection operation. All teacher models are assigned equal weights and we report the average accuracy.

\subsection{Analysis}
\subsubsection{Main Results}
Table~\ref{tab:main_results} compares three model scales across 11 benchmarks. At the 2B scale, SF-MOPD improves the domain-specific average from 38.4\% of the instruct baseline to 44.1\%, while maintaining general capability at 66.4\%. At the 4B scale, it improves the domain-specific average from 54.2\% of the instruct baseline to 56.9\%, outperforming MOPD by 0.5\%, while preserving general capability at 74.2\% and achieving the best overall average of 64.8\%. At the 8B scale, SF-MOPD reaches 60.4\% on the domain-specific average, an improvement of 3.2\% over the instruct baseline and 1.4\% over vanilla MOPD. General capability also improves to 76.4\%, a gain of 3.1\% over MOPD, narrowing the gap with the instruct baseline at 77.2\%. Our SF-MOPD approaches the Expert oracle reference on most domain-specific benchmarks.

\noindent\textbf{Effect of Capability Reference Design.}
To isolate the contribution of the reference mechanism, we compare two MOPD variants on the 8B model. MOPD with Ref adds a KL regularization term between the student and the fixed initialization model. While this improves general capability from 73.3\% to 73.8\% and domain-specific from 59.0\% to 59.1\% over vanilla MOPD, the fixed anchor constrains domain-specific learning, as the student is pulled toward a state without domain expertise. SF-MOPD replaces the fixed initialization with the EMA-smoothed student as a moving reference and adds the per-teacher slow-fast projection, achieving 60.4\% on domain-specific, a 1.4\% gain over vanilla MOPD and 1.3\% over MOPD with Ref, while improving general capability to 76.4\%, +3.1\% over MOPD and +2.6\% over MOPD with Ref. This confirms that the projection mechanism, rather than a fixed reference alone, drives the gains.

\begin{table}[htbp!]
\centering
\vspace{-0.1cm}
\caption{Comparison on general and domain-specific capability averages. The best result is shown in bold, and the second-best result is underlined.}
\label{tab:summary_results}
\vspace{-0.4cm}
\setlength{\tabcolsep}{3pt}
\renewcommand{\arraystretch}{1.0}
\small
\begin{tabular}{lccc}
\toprule
\textbf{Method} & \textbf{General} & \textbf{Specific} & \textbf{All} \\
\midrule
\rowcolor{gray!15} SF-MOPD (Ours) & \textbf{76.4} & \textbf{60.4} & \textbf{67.7} \\
SF-MOPD w/o. SFP & 72.3 & 57.0 & 64.0 \\
SF-MOPD w/o. $\mathcal{L}_{\mathrm{MOPD}}$ & \underline{75.9} & \underline{60.2} & \underline{67.3} \\
SF-MOPD w/o. CenterNorm & 75.4 & 59.7 & 66.8 \\
MOPD~\cite{ma2026mopd} & 73.3 & 59.0 & 65.5 \\
\bottomrule
\end{tabular}
\vspace{-0.8cm}
\end{table}

\subsection{Ablation Study}
As shown in Table~\ref{tab:summary_results}, we conduct ablation studies on the Qwen3-VL-8B model to evaluate our proposed components. The slow model, \emph{i.e.}, the EMA-ensembled student, serves as a reference. Removing Slow-Fast Projection~(SFP) reduces the full model to MOPD with an EMA-based reference constraint, where the training objective degenerates into directly pulling the student toward its EMA ensemble, \emph{i.e.}, the classic Mean-Teacher design. This variant drops the overall average from 67.7\% to 64.0\%, a decrease of 3.7\%. We then remove $\mathcal{L}_{\mathrm{MOPD}}$ and CenterNorm individually: dropping $\mathcal{L}_{\mathrm{MOPD}}$ slightly hurts both general and domain-specific capability, yielding 75.9\% General, 60.2\% Domain-Specific, and 67.3\% overall, suggesting that it further preserves the original optimization landscape and prevents catastrophic forgetting; removing CenterNorm also degrades performance to 66.8\% overall compared with 67.7\% for the full model, indicating it stabilizes the projection by aligning distributional statistics. The full SF-MOPD achieves 76.4\% General and 60.4\% Domain-Specific, reaching 67.7\% overall and outperforming MOPD at 65.5\% by 2.2\%. Among all components, SFP yields the largest individual gain with a 3.7\% drop when removed, underscoring the importance of the slow-fast projection mechanism.
We further conduct a sensitivity analysis on $\beta$: setting $\beta$ to $0.1$, $0.3$, $0.5$, $0.9$, and $1.0$ yields overall averages of $67.7\%$, $66.1\%$, $67.3\%$, $67.2\%$, and $67.0\%$, respectively.

\noindent\textbf{Conflict:}
Furthermore, we examined whether our method could yield additional gains when task heterogeneity conflicts are minimal. By increasing the sampling length of the fine-grained teacher training data, we reduced response heterogeneity across tasks; however, this strategy caused a severe degradation in fine-grained performance, with ZoomBench dropping to 40.7\%, indicating it is suboptimal for data construction. Comparing vanilla MOPD with SF-MOPD under these reduced-conflict settings, we found that general-purpose capability remained nearly unchanged at 76.5\% and 77.3\%, respectively. This suggests that even under minimal task heterogeneity conflicts, SF-MOPD still preserves general capabilities effectively.

\section{Conclusion}
We have presented SF-MOPD, a slow-fast multi-teacher on-policy distillation framework that addresses capability interference, where incorporating new domain expertise gradually erodes general capabilities. The fast model is the current student learning directly from each teacher, while the slow model, an EMA initialized from the same initialization, serves as a moving capability reference. The projection removes only the component of each teacher-induced update that pushes the fast model away from the slow one, leaving specialized learning uncapped. Extensive experiments across three model scales demonstrate that SF-MOPD consistently surpasses the instruct baseline and outperforms prior distillation methods, achieving the best domain-specific average at the 2B, 4B and 8B scales and improving the general-domain trade-off across scales. We hope this work encourages further research on balancing capability acquisition and preservation in MLLMs.

\section{ACKNOWLEDGMENT}
This work was supported by Ant Group Research Intern Program.

\section{Compliance with Ethical Standards}
All experiments in this work are purely computational (numerical simulations) and involve no human or animal subjects; therefore, no ethical approval was required.

\bibliographystyle{IEEEbib}
\bibliography{strings,refs}

\end{document}